\documentclass[11pt]{article}

\usepackage[preprint]{acl}

\usepackage{times}
\usepackage{latexsym}

\usepackage[T1]{fontenc}
\usepackage[utf8]{inputenc}

\usepackage{microtype}

\usepackage{inconsolata}

\usepackage{graphicx}
\usepackage{booktabs}
\usepackage{xspace}
\usepackage{amsmath}
\title{MOSAIC: A Multilingual, Taxonomy-Agnostic, and Computationally\\Efficient Approach for Radiological Report Classification}

\newcommand{\ku}{\textsuperscript{1}}
\newcommand{\nru}{\textsuperscript{2}}
\newcommand{\unumed}{\textsuperscript{3}}
\newcommand{\dr}{\textsuperscript{4}}
\newcommand{\cm}{\textsuperscript{5}}
\newcommand{\cerebriu}{\textsuperscript{6}}
\newcommand{\inns}{\textsuperscript{7}}
\newcommand{\comma}{\textsuperscript{,}}

\author{
    Alice Schiavone\ku\comma\nru,
    Marco Fraccaro\unumed,
    Lea Marie Pehrson\ku\comma\dr\comma\cm,
    Silvia Ingala\dr\comma\cerebriu,
    Rasmus Bonnevie\unumed
    \\
    \textbf{Michael Bachmann Nielsen}\cm,
    \textbf{Vincent Beliveau}\inns,
    \textbf{Melanie Ganz}\ku\comma\nru,
    \textbf{Desmond Elliott}\ku
    \\
    \ku Department of Computer Science, University of Copenhagen \\
    \nru Neurobiology Research Unit, Copenhagen University Hospital \\
    \unumed Unumed Aps,  
    \dr Department of Diagnostic Radiology, Copenhagen University Hospital \\
    \cm Department of Clinical Medicine, University of Copenhagen \\
    \cerebriu Cerebriu A/S,  
    \inns Institute for Human Genetics, Medical University of Innsbruck \\
    \texttt{alsc@di.ku.dk}, \texttt{de@di.ku.dk}
}

\usepackage{xcolor}
\usepackage{listings}
\usepackage{tcolorbox}

\definecolor{promptcolor}{RGB}{52, 154, 171} 
\definecolor{findingcolor}{RGB}{148, 220, 181} 
\definecolor{textcolor}{RGB}{26, 58, 94}

\lstdefinestyle{promptbox}{
  backgroundcolor=\color{promptcolor!15},
  frame=single,
  rulecolor=\color{promptcolor!55},
  framerule=1pt,
  framesep=8pt,
  linewidth=0.95\linewidth,  
  basicstyle=\ttfamily\small\color{textcolor},
  breaklines=true,
  breakatwhitespace=true,
  breakindent=0pt,
  columns=flexible,
  xleftmargin=10pt,
  xrightmargin=0pt,
  showstringspaces=false
}

\newcommand{\finding}[2]{%
  \begin{tcolorbox}[colback=findingcolor!20, colframe=findingcolor, arc=2pt, boxsep=0pt]
  \paragraph{\textbf{\textit{Finding} #1.}} #2
  \end{tcolorbox}%
}

\begin{document}

\maketitle
\begin{abstract}
Radiology reports contain rich clinical information that can be used to train imaging models without relying on costly manual annotation. However, existing approaches face critical limitations: rule-based methods struggle with linguistic variability, supervised models require large annotated datasets, and recent LLM-based systems depend on closed-source or resource-intensive models that are unsuitable for clinical use. Moreover, current solutions are largely restricted to English and single-modality, single-taxonomy datasets.
We introduce MOSAIC, a multilingual, taxonomy-agnostic, and computationally efficient approach for radiological report classification. Built on a compact open-access language model (MedGemma-4B), MOSAIC supports both zero-/few-shot prompting and lightweight fine-tuning, enabling deployment on consumer-grade GPUs. We evaluate MOSAIC across seven datasets in English, Spanish, French, and Danish, spanning multiple imaging modalities and label taxonomies. The model achieves a mean macro F1 score of 88 across five chest X-ray datasets, approaching or exceeding expert-level performance, while requiring only 24 GB of GPU memory. With data augmentation, as few as 80 annotated samples are sufficient to reach a weighted F1 score of 82 on Danish reports, compared to 86 with the full 1600-sample training set.
MOSAIC offers a practical alternative to large or proprietary LLMs in clinical settings. Code and models are open-source. We invite the community to evaluate and extend MOSAIC on new languages, taxonomies, and modalities.
\end{abstract}

\section{Introduction}

% Classic + Recent in English
Deep learning methods have been extensively explored for AI-assisted medical imaging analysis. However, their effectiveness depends on large volumes of annotated data. Such annotations must be provided by expert radiologists, whose primary focus remains clinical care, limiting the availability of high-quality labeled datasets. 
A promising solution to the annotation bottleneck is to extract relevant information directly from radiology reports, which are routinely produced during imaging procedures to document abnormalities associated with clinical findings \cite{Reichenpfader2024}. This information can later be used to train medical imaging classifiers or perform retrospective clinical studies \cite{Zhou2014Automated}.
Classic methods to automate finding extraction from English radiology reports include rule-based methods and BERT-based classifiers, that can prove effective in providing annotations close to those of radiologists \cite{irvin2019chexpert, chexbert}. However, rule sets need to be hand-crafted and are still limited by syntactic variability, and deep learning methods need a large amount of annotations from expert clinicians \cite{yang2023transformer}. Adapting these methods to a new label taxonomy or language necessitates either retraining the model, partially or entirely, or developing new rule sets from scratch. Re-training models to adapt to a new language or taxonomy for the same underlying task is an inefficient use of computational resources and is not sustainable.

In contrast, recent advances in natural language processing have led to the emergence of large language models (LLMs), which follow user instructions through natural language prompting. These models enable zero-shot or few-shot classification (i.e., with no or very few labeled examples), eliminating the need for manual annotation or rule engineering, and offering greater adaptability across tasks, languages, and taxonomies \cite{gu2024chex, dorfner2024open}. 
However, the use of LLMs in many clinical settings often depends on large and/or closed-source models that cannot be deployed locally, posing significant challenges for projects involving sensitive patient data due to privacy concerns and the need for high-end computational resources. In these scenarios, compact models that can operate on consumer-grade hardware are the most practical solution.

While several studies have explored this task, they are typically restricted to a single dataset, limited language coverage (often one or two languages), and limited to a single medical imaging modality, such as X-rays or magnetic resonance imaging (MRI), each having distinct clinical contexts and reporting conventions \cite{reis2022brax, nguyen2022learning, mottin2023multilingual, wollek2024german, italian2024, Matsuo2024, collado2025data, Mergen2025, ALMOHAMAD20252402}. Even when datasets share the same imaging modality, differences in research focus lead to variability in labeled findings. For example, a chest X-ray may be used to study either the heart or the lungs, resulting in highly diverse label sets across studies.
While LLMs offer greater flexibility than traditional deep learning methods, they still face important limitations. Most models are primarily trained on English, as it dominates the available web-crawled data. As a result, despite their strong performance on many tasks, LLMs underperform compared to BERT-based approaches when applied to other languages, e.g. Danish MRI reports \cite{beliveau2024classification}, or Japanese pancreatic cancer reports \cite{Suzuki2024}. 

To identify a viable approach to radiological report classification, we investigate whether generative language models can effectively perform this task in multiple languages and taxonomies, particularly in low-resource settings with limited computational capacity, scarce annotated data, and the need for localized model deployment. We look at language and task competency, evaluating performance across diverse linguistic settings and label taxonomies, and analyzing trade-offs between model size, accuracy, and adaptability.

We propose MOSAIC, an efficient and flexible LLM-based method for radiological report classification with the following key properties:

\begin{itemize}
    \item \textbf{Multilingual}: Trained and evaluated across multiple languages, including English, Spanish, French, and Danish.
    \item \textbf{Optimized for Small-scale}: Designed to be trained and tested on consumer-grade GPUs, enabling local model development and adaptation while preserving patient privacy by avoiding reliance on external services.
    \item \textbf{Adaptable}: Robust to variations in label taxonomies across datasets and medical imaging modalities. With data augmentation, the model can match full-dataset performance using as few as 80 examples.
    \item \textbf{Open-source}: We release the code and models in two sizes (4B and 12B)\footnote{Available upon paper acceptance}, while most datasets are accessible through their original providers.
\end{itemize}

We invite researchers to evaluate our method on their own data. Additional datasets provided by the community will be incorporated in a future revision of this manuscript.

\newcommand{\mimic}{\textsc{m}}
\newcommand{\reflacx}{$\textsc{r}$}
\newcommand{\reflacxi}{$\textsc{r}^\textsc{i}$}
\newcommand{\reflacxii}{$\textsc{r}^\textsc{ii}$}
\newcommand{\padchest}{\textsc{p$_{E+S}$}}
\newcommand{\padchestes}{\textsc{p$_S$}}
\newcommand{\padchesten}{\textsc{p$_E$}}
\newcommand{\casia}{\textsc{c}}
\newcommand{\danskmri}{\textsc{b}}
\newcommand{\danskcxr}{\textsc{d}}
\newcommand{\setone}{-11\xspace}
\newcommand{\settwo}{-25\xspace}

\newcommand{\mimictext}{\textit{MIMIC}}
\newcommand{\reflacxtext}{\textit{REFLACX}}
\newcommand{\reflacxitext}{\textit{REFLACX}$^I$}
\newcommand{\reflacxiitext}{\textit{REFLACX}$^{II}$}
\newcommand{\padchestestext}{\textit{PadChest}$_{es}$}
\newcommand{\padchestentext}{\textit{PadChest}$_{en}$}
\newcommand{\padchesttext}{\textit{PadChest}}
\newcommand{\casiatext}{\textit{CASIA}}
\newcommand{\danskcxrtext}{\textit{DanskCXR}}
\newcommand{\danskmritext}{\textit{DanskMRI}}

\newcommand{\llamaeight}{\texttt{Llama-8B}\xspace}
\newcommand{\llamathree}{\texttt{Llama-3B}\xspace}
\newcommand{\mmedllama}{\texttt{Mmed-\llamaeight}\xspace}
\newcommand{\gemmafour}{\texttt{Gemma-4B}\xspace}
\newcommand{\gemmatwelve}{\texttt{Gemma-12B}\xspace}
\newcommand{\medgemma}{\texttt{MedGemma-4B}\xspace}
\newcommand{\gemmatwentyseven}{\texttt{Gemma-27B}\xspace}
\newcommand{\medgemmatwentyseven}{\texttt{MedGemma-27B}\xspace}
\newcommand{\nllb}{\texttt{NLLB-3.3B}\xspace}
\newcommand{\llamaseventy}{\texttt{Llama-70B}\xspace}

\newcommand{\mosaic}{\textit{MOSAIC}\xspace}

\begin{table*}[t]
\centering
\small
\begin{tabular}{p{2.4cm} l c c c c c c c c}
\toprule
Dataset &  & Language & Modality & \begin{tabular}[c]{@{}c@{}}Number of\\Findings\end{tabular} & \begin{tabular}[c]{@{}c@{}}Avg.\\Chars\end{tabular} & \begin{tabular}[c]{@{}c@{}}Mention\\Classes\end{tabular} & Train & Dev & Test \\
\midrule
MIMIC-CXR & \mimic & en & Chest X-Ray & 14 & 760 & $+ , - ,\sim$ & 535 & 50 & 100\\
PadChest-GR & \textsc{p} & es, en & Chest X-Ray& 49 & 115 & $+$ & 1951 & 100 & 879\\
CASIA-CXR & \casia & fr & Chest X-Ray& 5 & 400 & $+$ & 7677 & 100 & 3334\\
DanskCXR & \danskcxr & da & Chest X-Ray & 48 & 312 & $+ ,-$ & 1600 & 125 & 750\\
Reflacx$^{I}$ & \reflacxi & en & Chest X-Ray & 14 & 216 & $+ ,\sim$ & 68 & 50 & 120 \\
Reflacx$^{II}$ & \reflacxii & en & Chest X-Ray & 15 & 201 & $+ ,\sim$ & 1046 & 52 & 1098\\
DanskMRI & \danskmri & da & Brain MRI & 3 & 1941 & $+ , - ,\sim$ & 194 & 50 & 345 \\
\bottomrule
\end{tabular}
\caption{Overview of the datasets used, including language, modality, number of findings, average characters per report, data split, and mention classes, namely positive \((+)\), negative \((-)\), and uncertain \((\sim)\) mentions of findings.}
\label{tab:datasets}
\end{table*}

\newcommand{\spacerule}{\addlinespace[0.5em]}

% \begin{table}[t]
% \small
% \centering
% \begin{tabular}{lp{4.6cm}}
% \toprule
% Label & Datasets \\
% \midrule
% \mimic & Mimic \\ \spacerule
% \mimic\padchesten & Mimic $\cup$ PadChest$_{en}$ \\  \spacerule
% \mimic\padchest & Mimic $\cup$ PadChest$_{en,es}$ \\ \spacerule
% \mimic\padchest\casia\danskcxr & Mimic $\cup$ PadChest$_{en,es}$ $\cup$ Casia $\cup$ DanskCXR \\ \spacerule
% %\mimic\padchest\casia\danskcxr \setone  & Mimic\setone $\cup$ PadChest$_{en}$\setone $\cup$ PadChest$_{es}$  $\cup$ Casia $\cup$ DanskCXR \\ \spacerule
% %\mimic\padchest\casia\danskcxr \settwo  & Mimic\settwo $\cup$ PadChest$_{en}$\settwo $\cup$ PadChest$_{es}$  $\cup$ Casia $\cup$ DanskCXR \\ \spacerule
% \mimic\padchest\casia$\Longrightarrow$\danskcxr & Model fine-tuned on \danskcxr $ $ from \mimic\padchest\casia $ $ checkpoint \\
% \bottomrule
% \end{tabular}
% \caption{.... We indicate with \setone a dataset that has been augmented by translating it into other 11 closely-related European languages. We indicate with \settwo a dataset that has been augmented by translating it into 25 European languages.}
% \label{tab:dataset-mapping}
% \end{table}

\section{MOSAIC}

\tcbset{
  inlinefindings/.style={
    boxsep=0.5mm,               % padding inside the box
    colback=green!10,           % light green background
    colframe=green!70!black,    % dark green border
    sharp corners=all,          % default, can change to rounded
    arc=2mm,                    % rounded corners
    boxrule=0.8pt,              % border thickness
    left=1mm, right=1mm, top=0.5mm, bottom=0.5mm,
    box align=base,             % align inline with text
    tcbox raise base          % vertical alignment
  }
}

MOSAIC is a language model specialized for prompt-based radiological report classification, available in either 4B and 12B versions. The final models are based on supervised fine tuning of the \medgemma and \gemmatwelve models using publicly available datasets. We describe our design decisions, including model selection, prompt design, fine-tuning strategies, and data augmentation to quantify the impact of each design choice.% presented as \tcbox[on line, colback=findingcolor!20, colframe=findingcolor, arc=2mm, boxrule=0.8pt, boxsep=0.2mm, left=0.8mm, right=0.8mm, top=0.2mm, bottom=0.2mm]{\textbf{\textit{Findings}}}.

% \begin{figure}[t]
%     \centering
%     %\includegraphics[width=0.8\linewidth]{imgs/firstpage.png}
%     \includegraphics[width=\linewidth]{imgs/figure1_cropped.pdf}
%     \caption{MOSAIC enables fine-tuning on multilingual, multi-taxonomy radiological reports using consumer-grade GPUs. It supports zero-/few-shot classification on new taxonomies, fine-tuning on new datasets and reports from different imaging modalities, and data augmentation of datasets with limited annotations.}
%     \label{fig:constructed-example}
% \end{figure}

\subsection{Data}

Few public radiological reports datasets are currently available, due to the risk of de-anonymization of patients or clinicians. Most reports are machine-annotated, which leads to noisy labels unsuitable for training language models. For high-quality results, we consider only datasets manually annotated or checked by radiologists. The datasets key statistics are available in Table~\ref{tab:datasets}.

\textit{MIMIC-CXR} \cite{johnson2019mimic} is a collection of chest X-ray English radiological reports. It has annotations for 3 possible types of mentioned finding: positive mention ($+$), when a finding is described as present in the report; negative mention ($-$), as finding is described as absent in the report; and uncertain mention ($\sim$), when a definitive conclusion cannot be reached on a specific finding. All not mentioned findings in the report are assigned a "not mentioned" label. Only a small subset of the dataset was manually annotated and released.

\textit{PadChest-GR} \cite{castro2024padchest} is a manually-curated chest X-ray dataset. The original dataset was a collection of reports from a hospital in Spain. Later, the sentences describing abnormalities were extracted from text through LLM prompting, and translated to English. The extracted sentences were reviewed by a team of radiologists while inspecting the associated X-ray. We limit the label set to findings that occur at least 30 times.  

\textit{CASIA-CXR} \cite{metmer2024open} is a French chest X-ray dataset. Each report was assigned one of five findings as positively mentioned. Labels for both \textit{PadChest-GR} and \textit{CASIA-CXR} can only be positively mentioned findings.

\textit{REFLACX} \cite{bigolin2022reflacx} is based on \textit{MIMIC-CXR}, from which unlabeled reports were extracted for manual annotation in two phases, with two different taxonomies: we refer to these as \textit{REFLACX$^{I}$} and \textit{REFLACX$^{II}$}. Each finding in these datasets was annotated as not mentioned, or as mention with varying degrees of certainty, with a score from 1 to 5, following the definition by \cite{panicek2016sure}. As granular uncertainty detection is not the scope of this study, we map probable findings as positive mentions (score of 4 and 5), and otherwise uncertain.

\textit{DanskCXR} \cite{Schiavone2025} is a private chest X-ray dataset collected from hospitals and clinics in Denmark. Along with \textit{MIMIC-CXR}, it is the only dataset that has annotations for negative mentions of findings. We limit our experiments to the 14 most frequent findings to exclude rare conditions.

We define as \textit{DanskMRI} a dataset of MRI reports from Danish hospitals collected by \citet{beliveau2024classification}, who focused on identifying epilepsy-related findings in brain MRI radiology reports. MRI is a different medical imaging modality: the style and structure of MRI reports differ significantly from those of chest X-rays, presenting a distinct out-of-distribution scenario. 

The dataset splits were generated using stratification \cite{sechidis2011stratification} and implemented using the \texttt{iterative-stratification} Python library, trying to have a consistent number of validation samples across datasets.

\subsection{Model selection and fine-tuning}

\begin{table*}[t]
\centering
\small
\begin{tabular}
{lp{0.25cm}p{0.25cm}p{0.4cm}p{0.25cm}p{0.25cm}p{0.4cm}p{0.25cm}p{0.25cm}p{0.4cm}p{0.25cm}p{0.25cm}p{0.4cm}p{0.25cm}p{0.25cm}p{0.6cm}p{0.25cm}p{0.25cm}p{0.25cm}p{0.25cm}}
\toprule
Dataset 
& \multicolumn{3}{c}{MIMIC} 
& \multicolumn{3}{c}{PadChest$_{en}$} 
& \multicolumn{3}{c}{PadChest$_{es}$} 
& \multicolumn{3}{c}{CASIA} 
& \multicolumn{3}{c}{DanskCXR} 
& \multicolumn{3}{c}{\textit{Average}} \\
\midrule
Experiment 
& ZS & 3S & FT 
& ZS & 3S & FT 
& ZS & 3S & FT 
& ZS & 3S & FT 
& ZS & 3S & FT 
& ZS & 3S & FT  \\
\midrule
\llamathree 
& 46 & 53 & 77 
& 60 & 66 & 67 
& 39 & 53 & 53 
& 55 & 75 & 79 
& 53 & 53 & 57 
& 50 & 60 & 66 \\
\llamaeight 
& 54 & 61 & 86 
& 77 & 76 & 78 
& 69 & 67 & 71 
& 70 & 75 & 77 
& 61 & 63 & 65 
& 66 & 68 & 75 \\
\mmedllama 
& 39 & 52 & 82 
& 42 & 75 & 78 
& 29 & 62 & 68 
& 60 & 67 & 77 
& 45 & 59 & 54 
& 43 & 63 & 72 \\
\gemmafour 
& 60 & 62 & 86 
& 68 & 75 & 74 
& 62 & 72 & 67 
& 69 & 73 & 80 
& 43 & 64 & 60 
& 60 & 69 & 73 \\
\medgemma 
& 55 & 59 & 88 
& 61 & 77 & 74 
& 53 & 72 & 71 
& 69 & 82 & 82 
& 62 & 65 & 58 
& 60 & 69 & 75 \\
\gemmatwelve 
& 65 & 70 & 84 
& 79 & 79 & 80 
& 76 & 76 & 76 
& 76 & 76 & 81 
& 69 & 75 & 69 
& 73 & 75 & 78 \\
\midrule
\gemmatwentyseven$*$ 
& 68 & 69 & 87 
& 81 & 81 & 82 
& 81 & 82 & 82 
& 75 & 77 & 81 
& 71 & 75 & 68 
& 75 & 77 & 80 \\
\medgemmatwentyseven$*$ 
& 70 & 69 & 87 
& 81 & 83 & 83 
& 81 & 84 & 82 
& 76 & 78 & 82 
& 72 & 76 & 70 
& 76 & 78 & 81 \\
\llamaseventy$*$ 
& 69 & 72 & 81 
& 78 & 79 & 78 
& 74 & 70 & 78 
& 68 & 79 & 79 
& 68 & 73 & 67 
& 71 & 75 & 77 \\
\bottomrule
\end{tabular}
\caption{Classification performance of language models on chest X-ray radiological free-text reports, measured in $(+)F1$ and ordered by model family and size. Models are tested under three settings: zero-shot (ZS); three-shot (3S) with three examples drawn from corresponding training sets; and the model fine-tuned on \mimictext\xspace (FT). We indicate with ($*$) models fine-tuned on a 94GB H100, instead of a consumer-grade 24GB RTX 4090.}
\label{tab:exp_direct_prompting}
\end{table*}

We select a small language model to base MOSAIC on through an initial set of experiments comparing two model families at different sizes: Llama 3 (3B and 8B) \cite{grattafiori2024Llama} and Gemma 3 (4B and 12B) \cite{team2025gemma} model families. These models have similar computational requirements and inherent multilingual capabilities. Additionally, we include \mmedllama, a multilingual medical foundation model  based on \llamaeight \cite{qiu2024towards}. Lastly, we test also three larger models of the same families (\gemmatwentyseven, \medgemmatwentyseven and \llamaseventy).  While these models are not feasible for local deployment on consumer-grade GPUs due to their high resource requirements, they serve to contextualize the performance of smaller alternatives. We select the instruction-tuned version of each model, given their general superiority for structured outputs \cite{zhang2025gpt4roi}. Closed-source models are excluded, as they do not meet the crucial privacy requirements of this task. 

Each model is fine-tuned on an NVIDIA RTX 4090 GPU (24GB) using its 4-bit quantized form along with Rank-Stabilized LoRA adapters, except for \gemmatwentyseven and \llamaseventy which are trained on a H100 (94GB). \mmedllama does not have a 4-bit quantized model version. All the prompts and findings sets are written in English to leverage the models’ stronger alignment to English instructions. The prompt structure is adapted based on the dataset's label taxonomy to ensure consistency with its definition of positive, uncertain, negative, and unmentioned findings. As an example, the prompt for \casiatext\xspace is:

\begin{lstlisting}[style=promptbox]
You are a helpful radiology assistant. Given a radiology report, classify each abnormality into a class. Output a valid JSON with each abnormality as key, and the class as value. The keys must be ['cardiomegaly', 'mass', 'pleural effusion', 'pneumonia', 'pneumothorax']. The values can be one of [-1, 1]. The values have the following interpretation: (1) the abnormality was mentioned, even with uncertainty, in the report e.g. 'A large pleural effusion', 'The cardiac contours are stable.', 'The cardiac size cannot be evaluated.';  (-1) the abnormality was not mentioned in the report, or the abnormality was negatively mentioned in the report; e.g. 'No pneumothorax.'.
\end{lstlisting}

Fine-tuning is conducted using the Unsloth library \cite{unsloth}, while inference is performed with vLLM \cite{kwon2023efficient}. Before inference, the LoRA adapters are merged into the base models in 16-bit precision. Cross-entropy loss is used as the objective function. 
Fine-tuning and sampling parameters are selected based on empirical testing and a random hyperparameter search conducted using Weights \& Biases \cite{wandb}. 
Training configurations, prompting strategy and hyperparameters are documented in the Appendix and accompanying code repository.

\subsection{Metrics}

We use the F1 score to evaluate the extraction of findings for positive and negative mentions. The average of these scores yields the macro F1 score, reported as $(+)F1$ and $(-)F1$, respectively. To mitigate the impact of class imbalance, we report the weighted F1 score as $(w)F1$, which incorporates class support into the macro F1 calculation.

\section{Results}

To evaluate a model's ability to classify the radiology reports across multiple languages and taxonomies, we look at two key aspects: language competency and task competency.

\subsection{Language competency}

Language competency is the ability of a model to understand and generate text in different languages. It reflects how the model captures linguistic patterns and vocabulary, often measured by metrics like perplexity. Strong language competency enables better performance across multilingual tasks.
SIB-200 is a large-scale, open-source benchmark for topic classification in 200 languages \cite{sib200}. To assess language modeling capabilities, Figure \ref{fig:exp_perplexity} shows the perplexity scores on SIB-200 across the four focus languages in this paper: English, French, Spanish, and Danish. 

Larger models such as \llamaeight and \gemmatwelve achieve lower perplexity compared to their smaller counterparts. The domain-adapted variants \mmedllama and \medgemma have different trends: while \medgemma outperforms its base model, \mmedllama does not. Contrary to expectations, we observe that model perplexity is often higher on English prompts compared to Spanish or French, despite English typically dominating model pretraining. 

\finding{1}{Larger language models reduce perplexity of general domain data, but domain-adapted variants show mixed results.}

%However, this might be explained by differences in tokenization efficiency. Perplexity is defined as $\text{PPL} = \exp\left(L/N\right)$ where $L$ is the total negative log-likelihood loss and $N$ is the number of tokens. In the SIB-200 dataset, English sentences average 26.5 tokens, while other languages average around 39 tokens. As English tokenizes more compactly, each token encapsulates more information, increasing the uncertainty (and loss) per token, thus leading to a higher perplexity. On English, this likely reflects greater information density per token rather than reduced model accuracy.

\begin{table*}[ht]
\centering
\small
\begin{tabular}{
p{1.5cm}
p{0.25cm}p{0.25cm}p{0.4cm}
p{0.25cm}p{0.25cm}p{0.4cm}
p{0.25cm}p{0.25cm}p{0.4cm}
p{0.25cm}p{0.25cm}p{0.4cm}
p{0.25cm}p{0.25cm}p{0.5cm}
p{0.25cm}p{0.25cm}p{0.8cm}
}
\toprule
 
& \multicolumn{3}{c}{MIMIC} 
& \multicolumn{3}{c}{PadChest$_{en}$} 
& \multicolumn{3}{c}{PadChest$_{es}$} 
& \multicolumn{3}{c}{CASIA} 
& \multicolumn{3}{c}{DanskCXR} 
& \multicolumn{3}{c}{\textit{Average}} \\
\midrule
Datasets
& \rotatebox{60}{\medgemma} & \rotatebox{60}{\llamaeight} & \rotatebox{60}{\gemmatwelve}
& \rotatebox{60}{\medgemma} & \rotatebox{60}{\llamaeight} & \rotatebox{60}{\gemmatwelve}
& \rotatebox{60}{\medgemma} & \rotatebox{60}{\llamaeight} & \rotatebox{60}{\gemmatwelve}
& \rotatebox{60}{\medgemma} & \rotatebox{60}{\llamaeight} & \rotatebox{60}{\gemmatwelve}
& \rotatebox{60}{\medgemma} & \rotatebox{60}{\llamaeight} & \rotatebox{60}{\gemmatwelve}
& \rotatebox{60}{\medgemma} & \rotatebox{60}{\llamaeight} & \rotatebox{60}{\gemmatwelve} \\
\midrule
\mimic  
& 87 & 86 & 84 
& 74 & 78 & 80 
& 71 & 71 & 76 
& 82 & 77 & 81 
& 58 & 65 & 69 
& 74 & 75 & 78 \\
\mimic\padchesten  
& 78 & \textbf{89} & 85 
& 92 & \textbf{95} & \textbf{95} 
& 84 & 85 & 91 
& 83 & 74 & 80 
& 65 & 65 & 70 
& 80 & 82 & 84 \\
\mimic\padchest  
& 80 & 82 & 84 
& 93 & 94 & \textbf{95} 
& 92 & \textbf{94} & \textbf{94} 
& 83 & 76 & 79 
& 61 & 61 & 69 
& 82 & 81 & 84 \\
\mimic\padchest\casia\xspace$\star$  
& 76 & 85 & 84 
& 92 & \textbf{95} & 94 
& 89 & \textbf{94} & \textbf{94} 
& \textbf{99} & \textbf{99} & \textbf{99} 
& 66 & 63 & 69 
& 84 & 87 & 88 \\
\mimic\padchest\casia\danskcxr  
& 76 & 76 & 85 
& 93 & 93 & \textbf{95} 
& 91 & 92 & \textbf{94} 
& \textbf{99} & \textbf{99} & \textbf{99} 
& 82 & 84 & \textbf{86} 
& 88 & 89 & 92 \\
\bottomrule
\end{tabular}
\caption{$(+)F1$ scores with incrementally expanded multilingual training configurations on  chest X-ray radiological reports. In bold, the best result over each dataset. \mimic\padchesten\xspace establishes a monolingual English baseline. Subsequent rows incrementally incorporate radiological reports in Spanish (\padchestes), French (\casia), and Danish (\danskcxr), culminating in a multilingual, multi-institutional training setup. \medgemma is 3 times smaller than \gemmatwelve, but it still achieves a competitive performance. The symbol $\star$ indicates the training configuration used for MOSAIC.}
\label{tab:exp_multilingual}
\end{table*}

\begin{figure*}[ht]
    \centering
    \includegraphics[width=0.9\linewidth]{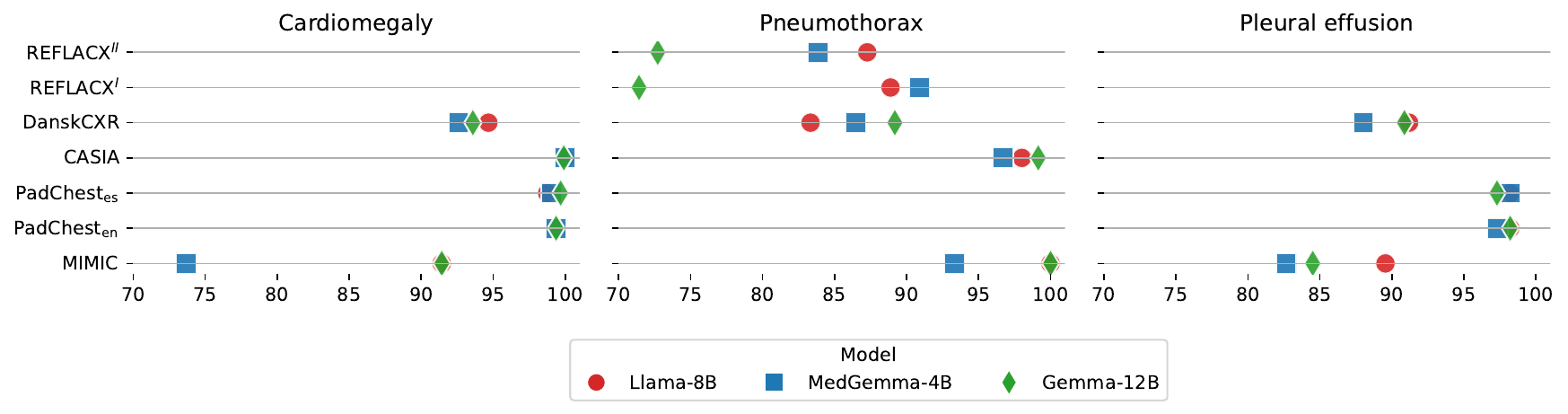}
    \caption{Performance as $(+)F1$ score of \medgemma, \llamaeight, and \gemmatwelve fine-tuned on \mimic\padchest\casia, with detailed results on three key Chest X-ray pathologies: Cardiomegaly, Pneumothorax, and Pleural Effusion. The data is presented across a range of chest X-ray datasets, illustrating model-specific and dataset-specific performance. On these common findings, all tested models have similar results, also generalizing on out-of-distribution datasets.}
    \label{fig:common-findings}
\end{figure*}

\subsection{Task competency}
\label{sec:task-competency}

Table~\ref{tab:exp_direct_prompting} reports the performance of each model across five chest X-ray datasets under three experimental settings, evaluated using the $(+)F1$ score: zero-shot (ZS), where the model is directly prompted to classify the text according to the given taxonomy; three-shot (3S), using three examples randomly drawn from the corresponding training sets; and fine-tuned (FT) on \mimictext. In the 3-shot setting, three examples from the respective training set are provided. 
When fine-tuning, the models were instructed to output a valid JSON with each abnormality as key, and the mention type as the value, using a zero-shot prompt. The outputs are then parsed and validated. Successful output rates for answers in valid JSON format are reported in the Appendix in Figure \ref{fig:invalid-answers}.

A clear trend emerges across all models and datasets: performance improves with 3-shot prompting. Even if fine-tuning is performed only on one dataset, some models achieve an improvement over the other tested datasets. \gemmatwelve consistently leads in performance, often achieving the highest F1 scores across multiple datasets and settings, particularly excelling in 3S and FT scenarios.
\medgemma also demonstrates strong performance, especially considering its smaller size, showing competitive F1 scores and even surpassing larger models in specific instances. 
\llamathree and \mmedllama generally show lower performance compared to Gemma variants, while \llamaeight performs generally well.
Surprisingly, \mmedllama performs poorly, especially compared to \llamaeight, particularly in zero-shot.
% compared to chexbert (sota)

On the \mimictext\xspace dataset, CheXbert reports radiologist-level performance with a $(w)F1$ of 0.809, compared to 0.743 for rule-based methods and 0.798 for BERT-based classifiers \cite{chexbert}. On a comparable subset, \medgemma\xspace achieves $(w)F1$=88, suggesting that our approach can achieve expert-level performance.

Large models (27B and 70B) have much higher scores on ZS and 3S, compared to the smaller models. When fine-tuned on the task, the performance is similar or equal on the target test set, but these models better generalize on the other test sets, unseen during fine-tuning. 

Figure \ref{fig:exp_prompting-violin} illustrates the distribution of the scores. While 0-shot learning shows a wide variability, the distributions become significantly tighter and higher for 3-shot learning. This trend is further pronounced when models are fine-tuned on \mimictext, suggesting a more robust and reliable performance. Interestingly, models consistently perform better on \padchestentext\xspace than on \padchestestext, despite both datasets containing the same reports but in different languages. This suggests that the task is easier in English than in Spanish.

We conclude that larger models offer a good solution when no annotated data is available, but lack the accuracy of fine-tuning on the target data. Remarkably, in the fine-tuned setting smaller models match their performance, but show reduced generalization to out-of-distribution datasets. Between the smaller models, \gemmatwelve exhibits superior performance but has higher memory and runtime demands. Additionally, training or inference may fail due to limited resources with large prompts (e.g. when the findings set or report is too long). %\llamaeight has a good balance between needed computing resource and accuracy, while \medgemma stands out for its efficiency and competitive performance. 

\finding{2}{In-context examples and supervised fine-tuning consistently improves performance compared to direct prompting. The best performing small-scale models are \medgemma and \gemmatwelve.}

\subsection{Multi-dataset fine-tuning}

Table~\ref{tab:exp_multilingual} reports $(+)F1$ scores using a series of increasingly comprehensive training setups, on the best-performing models in Section \ref{sec:task-competency}. Each successive row in the table reflects an incremental expansion of the training data, each indicated by the initial of the included dataset. The \mimic\padchesten\xspace setting includes only English-language data from \mimictext\xspace and \padchestentext, serving as the monolingual baseline. \mimic\padchest\xspace adds Spanish samples, introducing cross-lingual variation. In \mimic\padchest\casia, French data from CASIA is incorporated, followed by the final setting \mimic\padchest\casia\danskcxr, which adds Danish samples from DanskCXR, creating a multilingual, multi-institutional training configuration.
% language competency
We observe that the model trained under the \mimic\padchesten\xspace condition, using only English data, still achieves strong performance on \padchestestext, which contains the same reports as \padchesten\xspace in Spanish. This result suggests that the model can generalize to the task itself without direct exposure to Spanish during fine-tuning, providing evidence of emerging \emph{task competency}, even when the training data is exclusively monolingual, as long as the clinical structure remains consistent.

% task competency
While the inclusion of additional languages and datasets does lead to gradual performance improvements, these gains are most pronounced in the final configuration \mimic\padchest\casia\danskcxr, where all sources are present.  
The inclusion of \casiatext\xspace in \mimic\padchest\casia\xspace yields near-to-perfect results on \casiatext. This is not surprising, as this dataset is the largest (7.6k samples, compared to 0.5k of \mimictext) and with the simplest task, multi-class classification over 5 findings. 
Including \casiatext\xspace and \danskcxrtext\xspace slightly degrades performance on \mimictext\xspace in smaller models, likely due to \mimictext\xspace becoming less proportionally represented in the combined training data. This effect is not observed in \gemmatwelve. 

Figure \ref{fig:common-findings} further illustrates this trend for three critical pathologies, observed across the relevant training sets and the English-language datasets \reflacxitext\xspace and \reflacxiitext, under the \mimic\padchest\casia\danskcxr\xspace setting. While \gemmatwelve achieves the highest F1 scores, \medgemma performs on par with or better than larger models in several scenarios, and on the out-of-distribution \reflacxtext\xspace datasets.

\finding{3}{Adding more languages and datasets improves performance, with the largest gains in the full multilingual setting. English-only training transfers to Spanish due to shared clinical structure. }

\subsection{Taxonomy adaptation}

\newcommand{\checkemoji}{\includegraphics[width=1em]{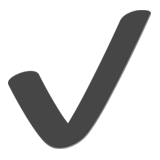}}
\newcommand{\crossemoji}{\includegraphics[width=0.7em]{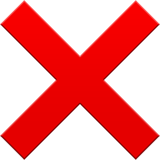}}

\begin{table}[]
\centering
\small
\begin{tabular}{cclcccccc}
\toprule
\mimic & \padchesten &  & \reflacxi & $\Rightarrow$\reflacxi & \reflacxii & $\Rightarrow$\reflacxii \\
\midrule
77 & 100 & Consolidation & 67 & 91 & 65 & 92 \\
93 & \crossemoji & Pneumothorax & 90 & 100 & 84 & 93 \\
\crossemoji & 68 & Nodule & 47 & 60 & \crossemoji & \crossemoji \\
\crossemoji & 100 & Hiatal Hernia & \crossemoji & \crossemoji & 89 & 100 \\
\crossemoji & \crossemoji & Emphysema & 71 & 50 & \crossemoji & \crossemoji \\
\crossemoji & \crossemoji & Enlarged Hilum & \crossemoji & \crossemoji & 74 & 75 \\
\bottomrule
\end{tabular}
\caption{Taxonomy adaptation in English of \mosaic trained on \mimic\padchest\casia\danskcxr, measured in $(+)F1$. Left columns show performance on present findings in the training sets \mimic\ and \padchesten; right columns show generalization to unseen datasets \reflacxi\ and \reflacxii\ before and after fine-tuning ($\Rightarrow$\reflacx). Per-class distribution shown in Table~\ref{tab:exp_taxonomy_support}.}
\label{tab:exp_taxonomy}
\end{table}

Table \ref{tab:exp_taxonomy} looks at the performance of \mosaic\xspace trained on \mimic\padchest\casia\danskcxr\xspace to assess task competency. The left columns (\mimic\xspace and \padchesten) reflect the model's initial task competency on English-language findings. The red "X" marks in these columns indicate that these specific findings are \textit{not present} in that dataset taxonomy. 
The right columns assess the model's task competency by measuring its ability to generalize to unseen English datasets \reflacxi and \reflacxii, both before and after fine-tuning on new data ($\Rightarrow$\textsc{r}). 
Although "Consolidation" is present in all datasets, fine-tuning on new data still yields substantial improvements. 
On findings that are present in only one of the training sets ("Nodule" and "Pneumothorax"), fine-tuning also significantly improves the models' ability to adapt their existing task competency to new distributions. 
However, for findings \textit{not} included in the initial training set taxonomy, the fine-tuned model shows a small improvement ("Enlarger Hilum", of which there are 29 samples in the whole dataset) or, in the case of "Emphysema" (N=6), fine-tuning hurts performance. 
%Our findings suggest that while fine-tuning aids in generalization, it does not fully compensate for data scarcity.

\finding{4}{Fine-tuning improves generalization to unseen datasets, especially for findings present in the training data, but offers limited gains for rare or missing findings.}

\subsection{Domain adaptation}

\begin{figure}
    \centering
    \includegraphics[width=\linewidth]{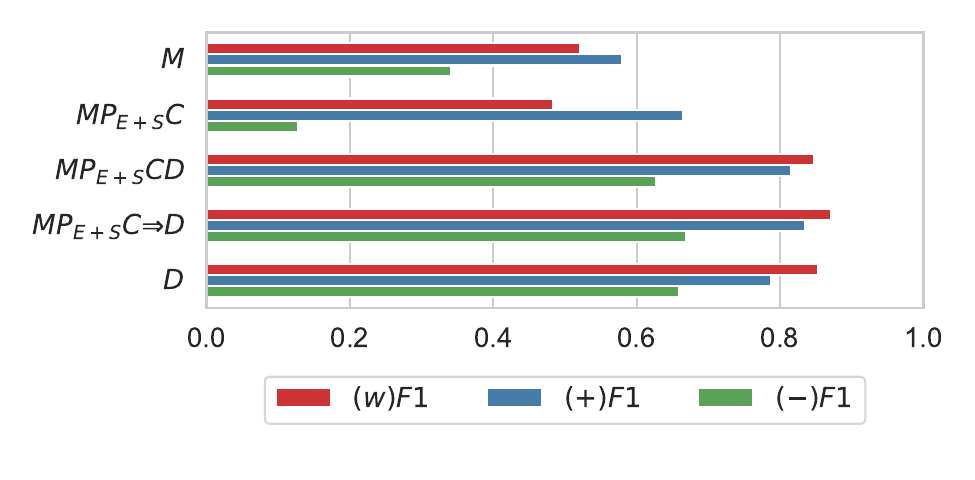}
    \caption{Evaluation of different pretraining and fine-tuning configurations of \mosaic on the \danskcxrtext. Fine-tuning directly on \danskcxr\xspace dataset outperforms zero-shot and task pretraining. %$(w)$F1=85, $(+)$F1=78. Sequential adaptation (\mimic\padchest\casia$\Rightarrow$\danskcxr) yields higher gains: $+2\%$ $(w)$F1, $+4\%$ $(+)$F1. 
    }
    \label{fig:exp_danskcxr_transfer}
\end{figure}

\danskcxrtext\xspace is challenging out-of-distribution (OOD) training set for evaluating model adaptation beyond in-domain performance. This dataset is complex, as it is multi-class, multi-label, sourced from multiple institutions, and written in Danish. Notably, the current SOTA on this task is a BERT-based model that achieves a $(+)F1$ score of 88, across both positively and negatively mentioned findings. 
Figure~\ref{fig:exp_danskcxr_transfer} shows how different training strategies impact performance on \danskcxrtext. When pretrained on \mimic\xspace or on \mimic\padchest\casia\xspace, the model struggles with negative finding detection:  $(-)F1$=34,  $(-)F1$=13, respectively. Incorporating \danskcxrtext\xspace into the pretraining phase yields no notable improvement over simply fine-tuning on \danskcxrtext\xspace alone. However, adapting a fine-tuned \mosaic\xspace to \danskcxrtext\xspace (\mimic\padchest\casia$\Rightarrow$\danskcxr) provides a modest performance boost, achieving $(+)F1$=84.

\begin{figure}
    \centering
    \includegraphics[width=0.8\linewidth]{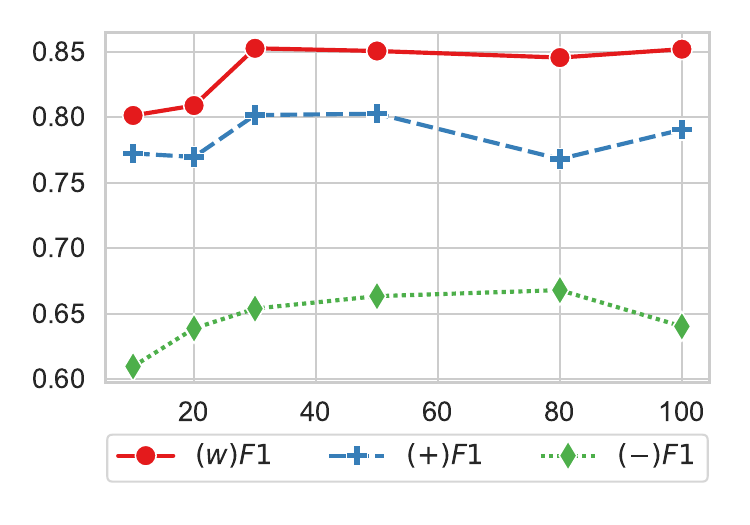}
    \caption{Data ablation study on \danskcxrtext\xspace shows that using just 480 examples (30\% of the full dataset) is sufficient to reach peak performance.}
    \label{fig:exp_danskcxr_abl}
\end{figure}

We next ask: what is the minimal data requirement to match full-dataset performance on \danskcxrtext? As shown in Figure~\ref{fig:exp_danskcxr_abl}, we find that using as little as 30\% of the dataset (480 examples) is sufficient to match the performance achieved with the full training set.
To further examine data efficiency of small LLMs, Table~\ref{tab:exp_danish_augmentation} presents results when training is restricted to just 5\% of the dataset (80 annotated examples). 

\finding{5}{Domain adaptation performance is maximized by sequential adaptation to the target domain dataset. Direct pretraining on mixed datasets offers limited gains.}

\finding{6}{Domain adaptation reaches its maximum performance with only 30\% of the expert-annotated data (480/1600 examples).}

\subsection{Data augmentation}

We also explore the benefit of data augmentation in domain adaptation. This experiment involves machine translating the original Danish dataset into English, obtained using \gemmatwentyseven, to which we refer to as \danskcxr$_e$. Models trained on this data were evaluated on the translated test set, which consistently yielded stronger results. Remarkably,  initializing \mosaic\xspace from \mimic\padchest\casia\xspace and fine-tuning on only 5\% of \danskcxr\xspace achieves a $(w)F1$ score just 4 points below that of the full-data setup, highlighting the impact of pretraining and the surprisingly low data requirement for effective adaptation with data augmentation.

\begin{table}[]
    \centering
    \small
    \begin{tabular}{llll}
    \toprule
     & 5\% & 10\% & 100\% \\
    \midrule
    \danskcxr & 75 & 80 & 85 \\
    \danskcxr$_{e}$ & 73 & 74 & 85 \\ 
    \danskcxr$_{e+d}$ & 79 & 80 & 86 \\ 
    \mimic\padchest\casia $\Rightarrow$ \danskcxr & 75 & 82 & 87 \\
    \mimic\padchest\casia $\Rightarrow$ \danskcxr$_{e+d}$ & 82 & 84 & 86 \\
    \bottomrule
    \end{tabular}

    \caption{$(w)F1$ scores of \mosaic trained on \danskcxrtext\xspace (\danskcxr) and its English translation (\danskcxr$_{e}$), with and without pretraining on \mimic\padchest\casia\xspace. Experiments are conducted on full data (100\%) and two training dataset subsets (5\% and 10\%).}
    \label{tab:exp_danish_augmentation}
\end{table}

\finding{7}{Data augmentation through machine translation improves classification performance when there are limited resources in a desired language. This can reduce the effort required for expert data annotation.}

\subsection{Imaging modality adaptation}

\begin{figure}
    \centering
    \includegraphics[width=0.8\linewidth]{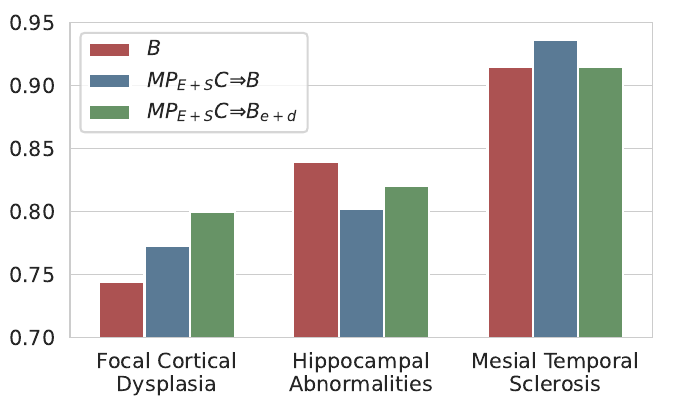}
    \caption{Performance of \mosaic on the \danskmritext\xspace dataset, measured as $(+)F1$ across three epilepsy-related abnormalities from MRI reports.}
    \label{fig:exp_danskmri}
\end{figure}

The \danskmritext\xspace dataset consists of Danish MRI reports annotated for three epilepsy-related brain abnormalities. Unlike chest X-ray datasets, these findings relate to neurological imaging, introducing both clinical and linguistic shifts.
As shown in Figure~\ref{fig:exp_danskmri}, adaptive fine-tuning on external chest X-ray datasets (\mimic\padchest\casia) improves performance over the base model for Focal Cortical Dysplasia and Mesial Temporal Sclerosis. This suggests that while cross-modality transfer can be effective, it may not generalize uniformly across all conditions.
Adding English data augmentation improves consistency across all findings. In particular, it recovers performance on Hippocampal Abnormalities without sacrificing gains on the others. These results highlight the benefit of lightweight augmentation when adapting to new modalities, especially under language and data constraints, as only 194 examples are provided for fine-tuning.

\finding{8}{Cross-imaging modality transfer improves performance; English data augmentation enhances consistency.}

\section{Conclusion}

We present MOSAIC, an approach for classifying radiology reports that prioritizes both local deployment and practicality. MOSAIC is a fine-tuned variant of a small language model, fine-tuned on publicly available multilingual chest X-ray datasets, including \mimictext, \padchesttext, and \casiatext.
Unlike existing approaches, MOSAIC is flexible across languages and label taxonomies, while remaining efficient enough to operate on standard consumer hardware. We evaluate MOSAIC on radiology reports in English, Spanish, French, and Danish across two imaging modalities and found that it performs robustly in all settings. When adapted to new data distributions with as few as 80 annotated examples, it delivers accurate predictions.

As the first method to unify efficient multilingual LLM adaptation with taxonomy-agnostic prompting, MOSAIC sets a new standard for radiology report classification, delivering state-of-the-art results across languages and taxonomies on affordable hardware. By eliminating reliance on large, expert-annotated datasets and language- or taxonomy-specific classifiers, MOSAIC offers a flexible, accessible alternative to prior methods. Its natural language prompting strategy ensures an optimal balance between memory efficiency and predictive accuracy, making it especially suitable for low-resource clinical environments.
MOSAIC is released in 4B and 12B parameter versions, both trained on publicly available datasets. We encourage the research community to explore, adapt, and extend MOSAIC, anticipating its contribution to responsible and scalable AI solutions in healthcare.

\section*{Limitations}

While MOSAIC shows strong performance across a range of datasets, it has several limitations. First, the model lacks interpretability: its predictions are based on internal transformer representations without transparent reasoning or token-level explanations. This opacity can be a barrier in clinical contexts, where trust and accountability are critical. Future work could explore integrating explainable components.
Second, for some datasets there are no established baselines, making it challenging to benchmark relative performance. Third, our evaluation focuses primarily on positively mentioned findings. However, some clinical applications may require precise handling of negative and uncertain statements, which we do not evaluate on.
Finally, although MOSAIC generalizes well across languages and imaging modalities, it struggles with rare or unseen conditions.

%\section*{Acknowledgments}

\bibliography{custom}

\appendix

\section{Appendix}
\label{sec:appendix}

\subsection{Label distribution}

\begin{figure}
    \centering
    \includegraphics[width=\linewidth]{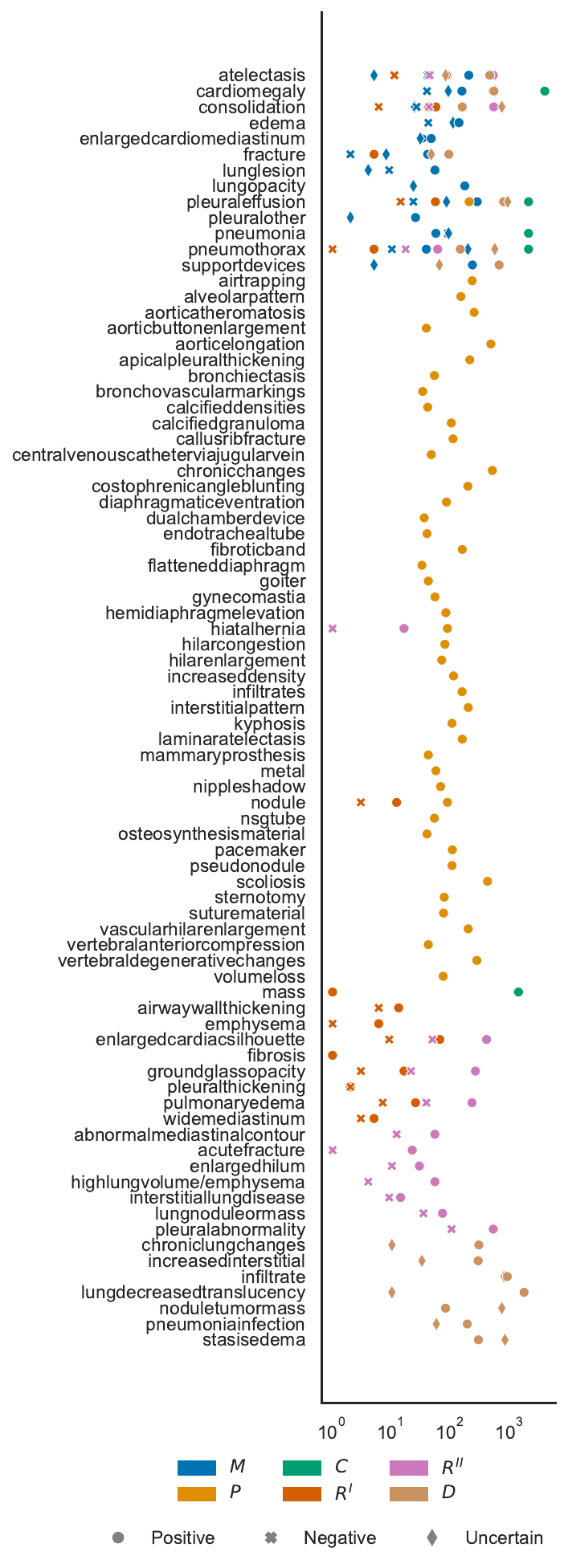}
    \caption{Distribution of labels in the chest X-ray datasets, by mention type and dataset.}
    \label{fig:taxonomy}
\end{figure}

Figure~\ref{fig:taxonomy} shows the distribution of labels in the six chest X-ray datasets used for this study. We distinguish each label by its mention type: positive, negative and uncertain. Many label are common across the datasets, but most are unique to a single dataset, particularly in the case of \padchesttext. This highlights the complexity of the task and emphasizes the importance of addressing label variety to ensure reliable model performance across different datasets. Negatively mentioned findings are also infrequently annotated, indicating a stronger focus on positively mentioned abnormalities.

\subsection{Prompt format}

Depending on the dataset taxonomy, we restructure the prompt accordingly. All prompts follow the expected structure for instruction-tuned models according to the models documentation. We start the prompt with a common header:
\begin{lstlisting}[style=promptbox]
You are a helpful radiology assistant. Given a radiology report, classify each abnormality into a class. Output a valid JSON with each abnormality as key, and the class as value. The keys must be {findings}. The values can be one of {classes}. The values have the following interpretation:
\end{lstlisting}
Then, if the dataset has uncertainly mentioned findings, we append to the prompt
\begin{lstlisting}[style=promptbox]
(1) the abnormality was positively mentioned in the report; 
\end{lstlisting}
Otherwise
\begin{lstlisting}[style=promptbox]
(1) the abnormality was mentioned, even with uncertainty, in the report, e.g. 'A large pleural effusion', 'The cardiac contours are stable.', 'The cardiac size cannot be evaluated.';
\end{lstlisting}
If negatively mentioned findings are included, we append
\begin{lstlisting}[style=promptbox]
(2) the abnormality was negatively mentioned in the report; e.g. 'No pneumothorax.'
\end{lstlisting}
If the dataset has uncertainly mentioned findings
\begin{lstlisting}[style=promptbox]
(0) the abnormality was either: mentioned with uncertainty in the report,or mentioned with ambiguous language in the report and it is unclear if the pathology exists or not, e.g. Explicit uncertainty: 'The cardiac size cannot be evaluated.', Ambiguous language: 'The cardiac contours are stable.'
\end{lstlisting}
Finally, we take care of unmentioned findings if negated findings are present as
\begin{lstlisting}[style=promptbox]
(-1) the abnormality was not mentioned in the report.
\end{lstlisting}
Otherwise
\begin{lstlisting}[style=promptbox]
(-1) the abnormality was not mentioned in the report, or the abnormality was negatively mentioned in the report; e.g. 'No pneumothorax.'
\end{lstlisting}
The listed examples are from the task definition of \mimictext.

\subsection{Model Performance Across Languages and Tasks}

\begin{figure}
    \centering
    \includegraphics[width=0.9\linewidth]{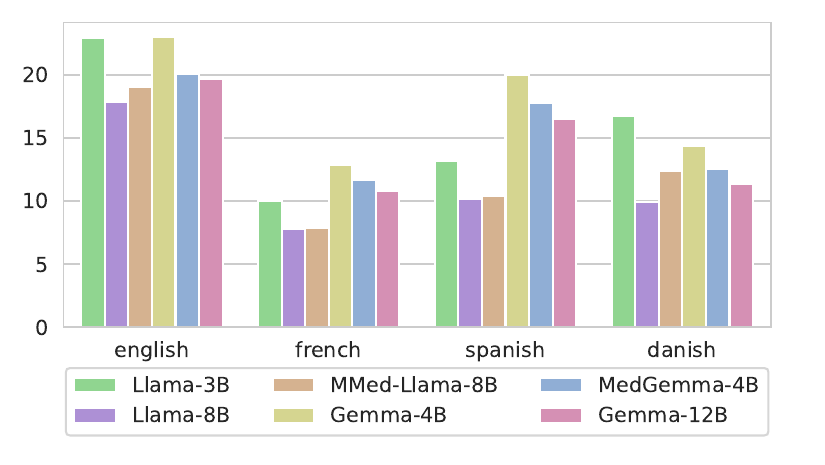}
    \caption{Perplexity on the SIB-200 dataset across English, French, Spanish, and Danish. Lower values indicate better language modeling. \llamaeight and \gemmatwelve achieve the lowest perplexities overall, while medical-domain variants (\mmedllama , \medgemma  ) perform competitively, especially in French and Spanish.}
    \label{fig:exp_perplexity}
\end{figure}

\begin{figure}
    \centering
    \includegraphics[width=0.8\linewidth]{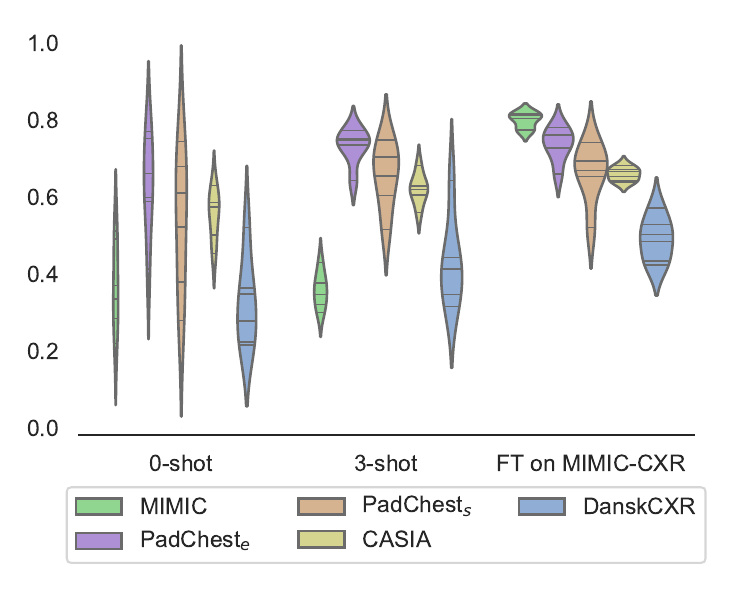}
    \caption{Distribution of $(+)F1$ scores from Table \ref{tab:exp_direct_prompting}. Performance improves with increased supervision, but transferability varies by dataset. Fine-tuning on \mimictext\xspace yields higher and more consistent scores on all datasets.}
    \label{fig:exp_prompting-violin}
\end{figure}

Figures~\ref{fig:exp_perplexity} and \ref{fig:exp_prompting-violin} provide complementary evaluations of language and task competency. Figure~\ref{fig:exp_perplexity} shows model perplexity on SIB-200, indicating that larger and medical-domain models achieve lower perplexity across languages. Figure~\ref{fig:exp_prompting-violin} illustrates few-shot and fine-tuned performance, highlighting that additional supervision improves $(+)F1$ scores and yields more consistent results across datasets.

\subsection{Instruction following}

\begin{figure*}
    \centering
    \includegraphics[width=0.9\textwidth]{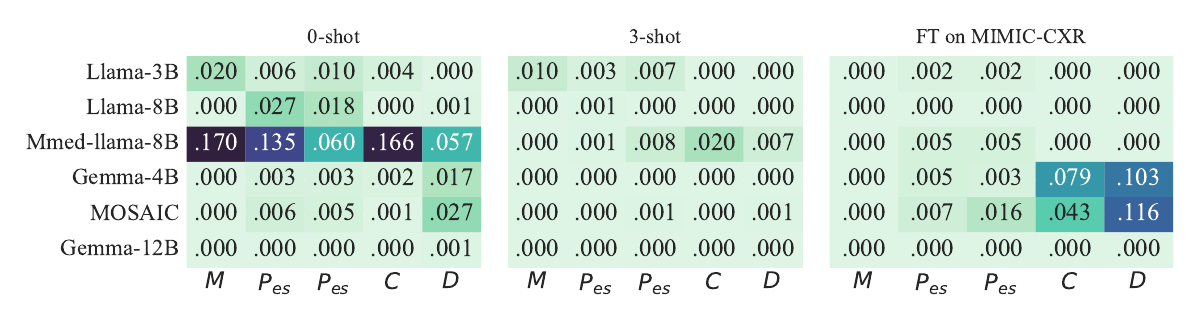}
    \caption{Percentage of invalid outputs when testing each model setting in Table \ref{tab:exp_direct_prompting}. For an answer to be considered valid, it must be correctly parsed as a JSON file. This is achieved using regular expressions and a safe evaluation method to convert strings into Python objects. Answers may still be marked invalid if the classification does not follow the expected format. Missing findings are supplemented with a default value, typically "finding not mentioned."}
    \label{fig:invalid-answers}
\end{figure*}

Figure~\ref{fig:invalid-answers} presents the percentage of invalid outputs produced by each model under 0-shot, 3-shot, and fine-tuned (FT) settings across five test sets. Invalid outputs are those that fail to meet the expected JSON structure. Overall, fine-tuning results in consistently well-structured outputs. Zero-shot prompting leads to a higher rate of formatting errors, especially for \mmedllama. \gemmafour and \medgemma exhibit more issues in the FT setting, suggesting overfitting and issues to adhere the given instruction on unseen datasets.

\begin{table}[]
    \centering
    \small
    \begin{tabular}{lrr}
    \toprule
     & Memory use& Runtime \\
    \midrule
    \llamathree & 4.70 & 9.66 \\
    \mosaic & 8.69 & 13.98 \\
    \gemmafour & 9.16 & 13.88 \\
    \llamaeight & 9.61 & 20.68 \\
    \gemmatwelve & 16.24 & 33.64 \\
    \mmedllama & 19.01 & 19.33 \\
    \midrule
    \gemmatwentyseven & 43.85 & 36.58 \\
    \medgemmatwentyseven & 41.48 & 34.65 \\
    \llamaseventy & 78.14 & 79.27 \\
    \bottomrule
    \end{tabular}
    \caption{Training runtime (in minutes) and peak GPU memory usage (in GB) for models trained on \mimictext\xspace. Training is performed using LoRA on 4-bit quantized models, except for \mmedllama. \medgemma, MOSAIC's base model, uses 9.75}
    \label{tab:exp_memory_use}
\end{table}

\subsection{Model Size Comparison on \danskcxrtext}

\begin{figure}[ht]
\centering
\includegraphics[width=0.8\linewidth]{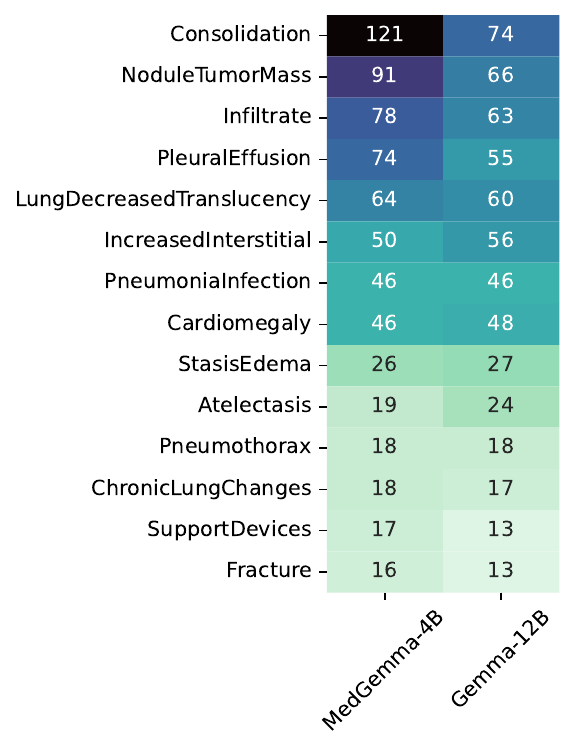}
\caption{Prediction mismatches on \texttt{DanskCXR}: \medgemma\ (684 mismatches; 6.5\%) vs. \gemmatwelve\ (580 mismatches; 5.5\%).}
\label{fig:exp_model_error_comparison}
\end{figure}

We compare two MOSAIC variants based on different underlying model sizes: \medgemma (4B) and \gemmatwelve (12B). On the \texttt{DanskCXR} dataset, \gemmatwelve outperforms \medgemma with a weighted F1 score of 88 [$(+)F1$=86, $(-)F1$=0.67)], compared to \medgemma’s score of 85 [$(+)F1$=82, $(-)F1$=0.63]. This performance gap is reflected in the overall number of prediction mismatches: 684 for \medgemma versus 580 for \gemmatwelve, as illustrated in Figure~\ref{fig:exp_model_error_comparison}.

A deeper inspection reveals that many of \medgemma's errors stem from over-prediction of certain conditions, particularly consolidation. This tendency is illustrated by the following representative example:

\begin{quote}
\textit{Chest X-ray (RU of thorax) taken in two planes, compared with [date], shows no pneumothorax or pleural effusion. Stationary conditions are observed with rounded infiltrates bilaterally. No signs of congestion or atelectasis. Slim mediastinum and normal heart size.}
(Translated from Danish)
\end{quote}

Classification comparison, where (-) is the "not mentioned" class:

\begin{center}
\small
\begin{tabular}{lccc}
\toprule
Finding & \rotatebox{60}{Ground Truth} &
\rotatebox{60}{\gemmatwelve}&
\rotatebox{60}{\medgemma} \\
\midrule
Atelectasis & 2 & 2 & 2 \\
Cardiomegaly & 2 & 2 & 2 \\
Infiltrate & 1 & 1 & 1 \\
LungDecr.Transl. & 1 & 1 & 1 \\
PleuralEffusion & 2 & 2 & 2 \\
Pneumothorax & 2 & 2 & 2 \\
StasisEdema & 2 & 2 & 2 \\
Consolidation & - & - & \textbf{1} \\
\bottomrule
\end{tabular}
\end{center}

In this case, \medgemma\ incorrectly predicts consolidation. Both models agree with the ground truth for the remaining findings, reinforcing the observation that \gemmatwelve’s larger capacity yields modest yet consistent gains in precision—particularly for subtle or infrequent findings.

However, the larger model is not immune to errors. A notable case follows:

\begin{quote}
\textit{X-ray examination of the chest in two projections, standing, on [date] with no previous images for comparison. Shows normal conditions of the heart and lungs.}
(Translated from Danish)
\end{quote}

Here, \gemmatwelve\ incorrectly labels Cardiomegaly as a negative mention, despite the report explicitly describing normal heart size. Interestingly, \medgemma avoids this error. While this misclassification may be due to an ambiguity in phrasing or even in the labeling protocol rather than a fundamental model limitation, it highlights that increased model capacity does not universally guarantee better performance across all contexts or findings.

\subsection{Training details}

All models were trained for up to 5 epochs using the AdamW optimizer with 8-bit precision and a cosine learning rate schedule. We used a learning rate of 1e-4, a warmup ratio of 0.05, and weight decay of 0.001 for Llama models, and of 0.01 for Gemma models. Training was conducted with a target device batch size of 64, which was achieved through a number of gradient accumulation steps depending on the model size. Early stopping was applied with a patience of 5 evaluation steps and a threshold of 0.001. Models were evaluated every 10 steps if a single dataset was used for fine-tuning, or 30 evaluatation steps if the experiment included more than one dataset. Mixed precision training was enabled using bfloat16 (bf16), and a fixed random seed (42) was used for reproducibility. Models were loaded in 4-bit precision to reduce memory usage. For parameter-efficient fine-tuning, we used Rank-Stabilized LoRA with a rank of 64 and scaling factor of 128, and no dropout. LoRA adapters were injected into key projection layers within the transformer blocks, as well as the MLP layers. No additional bias parameters were trained. Gemma vision-language models were trained using only their text encoder

We conducted inference using the vLLM engine with optimized settings to balance efficiency and output quality. GPU memory utilization was capped at 90\% to ensure stable performance, with a maximum sequence length of 2048 tokens (doubled for 3-shot settings) and support for up to 64 concurrent sequences. The ability to process long sequences has previously been shown to substantially improve performance on the MIMC-III dataset~\cite{dai-etal-2022-revisiting}. LoRA adaptation was enabled with a maximum rank of 64.

For generation, we used a temperature of 0.5 to encourage output diversity while maintaining coherence. A minimum probability threshold (min\_p) of 0.1, and generation was configured to stop after encountering a closing brace (\}), as we require a JSON output. To ensure reproducibility, a fixed random seed was used (42).

\subsection{Class support in English taxonomies}

\begin{table}[]
\centering
\small
\begin{tabular}{c c l l c}
\toprule
\mimic & \padchesten & Finding &  & S \\
\midrule
7 & 8 & Consolidation & \reflacxi & 32 \\
  &   &               & \reflacxii & 260 \\
7 & \crossemoji & Pneumothorax & \reflacxi & 5 \\
  &             &              & \reflacxii & 31 \\
\crossemoji & 19 & Nodule & \reflacxi & 4 \\
\crossemoji & 27 & Hiatal Hernia & \reflacxii & 9 \\
\crossemoji & \crossemoji & Emphysema & \reflacxi & 2 \\
\crossemoji & \crossemoji & Enlarged Hilum & \reflacxii & 11 \\
\bottomrule
\end{tabular}
\caption{Class support in each test set for the taxonomy adaptation in English experiment on \mimic\ and \padchesten (Table~\ref{tab:exp_taxonomy}).}
\label{tab:exp_taxonomy_support}
\end{table}

In Table~\ref{tab:exp_taxonomy_support} we list the support for each finding in the target test set. "Emphysema" and "Enlarged Hilum" are not present in the training set of MOSAIC, and are a rare finding also in \reflacxitext and \reflacxiitext. 

\subsection{Data Augmentation}

\begin{table}
\small
\centering
\begin{tabular}{lcc}
\toprule
 & $EN \rightarrow ES$ &  $ES \rightarrow EN$ \\
\midrule
\nllb & 63 & 55 \\
\llamathree & 61 & 66 \\
\llamaeight & 68 & 65 \\
\mmedllama & 42 & 57 \\
\medgemma   & 69 & 76 \\
\gemmafour  & 72 & 74 \\
\gemmatwelve& 76 & 80 \\
\gemmatwentyseven & 82 & 81 \\
\bottomrule
\end{tabular}
\caption{Translation results for the models in the paper and fb-nllb, on PadChest$_{en,es}$. The table shows the METEOR score for each model and language direction.}
\label{tab:translation_padchest}
\end{table}

To enhance cross-language performance, we investigate machine translation as a data augmentation strategy. We evaluate the translation quality of our selected models alongside \nllb, a large language model specifically trained for machine translation across over 200 languages. The prompt used was

\begin{lstlisting}[style=promptbox]
    Translate this text into {language}. Respond only with the translation.
\end{lstlisting}

Some translation could exceed the maximum length allowed during training, so these were cut from the dataset. For the evaluation of machine translation quality for data augmentation, we employ the METEOR metric \cite{banerjee2005meteor}, as it is considered as a good proxy for human-like translation assessment.

We begin by examining \textit{PadChest}. Despite its shorter report lengths compared to other datasets, its availability in both English and Spanish makes it valuable for evaluating bidirectional translation. The results presented in Table \ref{tab:translation_padchest} highlight \gemmatwentyseven as the top-performing model. In contrast, \nllb and \mmedllama show lower translation quality on this task.

To compare the performance of the Llama and Gemma model families on Danish, we evaluated their Danish-English backtranslation capabilities. According to their official documentation, \llamaeight was not explicitly trained on Danish. \gemmatwentyseven achieved a METEOR score of 0.58, significantly outperforming \llamaeight, which scored 0.45. Consequently, we selected \gemmatwentyseven's translations for our data augmentation experiments. This model it suitable for inference on our hardware, while its memory demands exceed our training capacity.

\end{document}